\documentclass[11pt]{article} 
\usepackage[utf8]{inputenc}

\usepackage[utf8]{inputenc} 
\usepackage[T1]{fontenc}    
\usepackage{hyperref}       
\usepackage{url}            
\usepackage{booktabs}       
\usepackage{amsfonts}       
\usepackage{nicefrac}       
\usepackage{microtype}      

\usepackage[space]{grffile}
\usepackage{amsmath,amsfonts,amssymb}
\usepackage{mdframed}
\usepackage{mathtools}
\usepackage{bbm}
\usepackage{mathrsfs}
\usepackage{bm}
\usepackage{dsfont}
\usepackage{latexsym}
\usepackage{graphicx}
\usepackage{fancyvrb}
\usepackage{nicefrac}
\usepackage{enumerate}
\usepackage{dashrule}
\usepackage{enumitem}

\usepackage[most]{tcolorbox}
\usepackage{parskip}
\usepackage{xcolor,colortbl}

\usepackage[square, sort, numbers]{natbib}
\newcommand{\set}[1]{\left\{#1\right\}}			
\newcommand{\abs}[1]{\left|#1\right|}
\newcommand{\paren}[1]{\left(#1\right)}
\renewcommand{\brack}[1]{\left[#1\right]}
\newcommand{\norm}[1]{\left\lVert #1 \right\rVert}	

\newcommand{\st}{\text{ s.t. }}

\renewcommand{\wp}{\text{ w.p. }}
\newcommand{\R}{\mathbb{R}}         	
\newcommand{\E}{\mathbf{E}}
\renewcommand{\P}{\mathbb{P}}

\newcommand{\given}{\;\middle\vert\;}

\newcommand{\one}[1]{\,\mathbf{1}{\left\{{#1}\right\}}}	

\newcommand{\ppv}{\text{ppv}}
\newcommand{\npv}{\text{npv}}
\newcommand{\tpr}{\text{tpr}}
\newcommand{\tnr}{\text{tnr}}

\renewcommand{\r}[1]{\boldsymbol{#1}}

\renewcommand{\d}[1]{\,\text{d}{#1}\;}

\newcommand*{\qed}{\hfill\ensuremath{\square}} 

\DeclareMathSymbol{@}{\mathord}{letters}{"3B} 

\DeclareMathOperator*{\argmax}{arg\,max}

\allowdisplaybreaks

\begin{document}
\thispagestyle{empty}
\centerline{\bf \LARGE Predictive Value Generalization Bounds}
\vspace{1mm}
\noindent\hrulefill
\vspace{.5 cm}

\textbf{Keshav Vemuri}, \textit{Department of Statistics, University of Chicago} \hfill \texttt{kvemuri@uchicago.edu}

\textbf{Nathan Srebro}, \textit{Toyota Technological Institute at Chicago} \hfill \texttt{nati@ttic.edu}

\vspace{.5 cm}

{
\small
\begin{center}
\subsection*{Abstract}
\end{center}

\begin{quotation}
\noindent
In this paper, we study a bi-criterion framework for assessing scoring functions in the context of binary classification. The positive and negative predictive values (ppv and npv, respectively) are conditional probabilities of the true label matching a classifier's predicted label. The usual classification error rate is a linear combination of these probabilities, and therefore, concentration inequalities for the error rate do not yield confidence intervals for the two separate predictive values. We study generalization properties of scoring functions with respect to predictive values by deriving new distribution-free large deviation and uniform convergence bounds. The latter bound is stated in terms of a measure of function class complexity that we call the order coefficient; we relate this combinatorial quantity to the VC-subgraph dimension. 
\end{quotation}
}

\section{Introduction}

Binary classification is a fundamental problem that has received enormous attention from theoretical and applied communities for many decades. Any particular choice of classifier produces a $2\times 2$ \textit{confusion matrix} that tabulates its performance on labeled training data, and from which many useful statistics can be derived (see Figure \ref{fig:1}). In this paper, we study the predictive accuracy of threshold-based classifiers via two empirical probabilities derived from the confusion matrix: \textit{positive predictive value} (ppv), the proportion of positively classified training examples whose true label is positive, and \textit{negative predictive value} (npv), the proportion of negatively classified examples whose true label is negative. If the pair $(\r{x}, \r{y})\in \mathcal{X}\times \set{0,1}$ 
denotes a randomly selected labeled example, and we use the threshold classifier $h(x) = \one{f(x) > t}$ for some scoring function $f:\mathcal{X}\rightarrow \R$, the population counterparts of the empirical predictive values can be denoted as
\[
\ppv(t) = \P\set{\r{y} = 1 \given f(\r{x}) > t}\quad, \quad \npv(t) = \P\set{\r{y} = 0 \given f(\r{x}) \le t}\;.
\] 
As discussed in prior works \cite{mp04, sg08}, these quantities provide the type of information needed for clinical decision-making. For example, if $[.98, 1]$ is a $95\%$ confidence interval for $\npv$ and $[.37, .57]$ is a $95\%$ confidence interval for $\ppv$, then negative diagnoses made by the classifier are likely to be accurate, while positive diagnoses are more likely spurious and will require further testing. 

In sharp contrast with the parametric modeling approaches taken in the papers cited above, we will analyze predictive accuracy in the distribution-free framework of statistical learning theory. We treat two distinct settings: one in which the scoring function $f$ is known, and one in which $f$ is unknown and must be selected from a class of functions $\mathcal{F}$. \textit{In the first setting, we prove a large deviation bound to construct finite-sample, distribution-free confidence bands for the predictive value curves $\ppv(f, \cdot)$ and $\npv(f, \cdot)$, which trace out the ppv and npv of $f$ over all threshold levels. In the second setting, where the choice of scoring function may depend on the same data used to evaluate its performance, we prove a uniform convergence bound and establish confidence bands for the predictive value curves that hold uniformly over the class $\mathcal{F}$. }

The predictive value curves here are reminiscent of receiver operating characteristic (ROC) curves and precision-recall (PR) curves, which are commonly used graphical tools in a variety of applications \cite{sr15, ob18, gsob18, dg06, h19}. For a fixed scoring function $f$, the ROC curve consists of pairs $(\text{tpr}(t), 1 - \text{tnr}(t))$ for different threshold settings $t$, where tpr is the true positive rate, $\P\set{f(\r{x}) > t \given \r{y} = 1}$, and tnr is the true negative rate, $\P\set{f(\r{x}) \le t \given \r{y} = 0}$. There is a long history of statistical research on the topic of parametric and nonparametric estimation of ROC curves \cite{ht96, hhf04, jp13}. However, despite their popularity in evaluating scoring functions, true positive rate (tpr), $\P\set{f(\r{x}) > t \given \r{y} = 1}$, and true negative rate (tnr), $\P\set{f(\r{x}) \le t \given \r{y} = 0}$, can provide a misleadingly optimistic view of classification performance when the label bias, $\P\set{\r{y}=1}$, is close to 0 or 1 (a situation commonly referred to as \textit{class imbalance}). As seen in Figure \ref{fig:1}, the pair of column ratios, $(\tpr, \tnr) = (.9, .9)$, remains unchanged across the two datasets, whereas the pair of row ratios, $(\ppv, \npv)$, shifts from $(.9, .9)$ to $(.47, .99)$ when moving from the balanced to the imbalanced dataset; in medical screening applications, imbalanced datasets are very common, and it's clear that a classifier with ppv of $.47$ will be unable to accurately predict positive disease status. 

\begin{figure}[ht]
\begin{center}
\begin{tabular}{|l|l|l|l|}
\hline
\cellcolor[HTML]{EFEFEF} & True $+$                   & True $-$                   & \cellcolor[HTML]{EFEFEF} \\ \hline
Predicted $+$            & \cellcolor[HTML]{34FF34}90 & \cellcolor[HTML]{FD6864}10 & 100                      \\ \hline
Predicted $-$            & \cellcolor[HTML]{FD6864}10 & \cellcolor[HTML]{34FF34}90 & 100                      \\ \hline
\cellcolor[HTML]{EFEFEF} & 100                        & 100                        & \cellcolor[HTML]{EFEFEF} \\ \hline
\end{tabular}
\hspace{1cm}
\begin{tabular}{|l|l|l|l|}
\hline
\cellcolor[HTML]{EFEFEF} & True $+$                   & True $-$                    & \cellcolor[HTML]{EFEFEF} \\ \hline
Predicted $+$            & \cellcolor[HTML]{34FF34}90 & \cellcolor[HTML]{FD6864}100 & 190                      \\ \hline
Predicted $-$            & \cellcolor[HTML]{FD6864}10 & \cellcolor[HTML]{34FF34}900 & 910                      \\ \hline
\cellcolor[HTML]{EFEFEF} & 100                        & 1000                        & \cellcolor[HTML]{EFEFEF} \\ \hline
\end{tabular}
\end{center}
\caption{Numerical examples, as shown by \citet{h19}, of confusion matrices for a threshold-based classifier $h(x) = \one{f(x) > t}$ evaluated on a balanced sample (left) and an imbalanced sample (right).}
\label{fig:1}
\end{figure}

The Precision-Recall (PR) curve is much closer in spirit to the objects we are interested in and consists of the points $(\text{ppv}(t), \text{tpr}(t))$ as $t$ varies. This construction is very similar to the re-parameterized ppv curve that we will later analyze in detail, which plots ppv, $\P\set{\r{y} = 1 \given f(\r{x}) > t}$, against the \textit{positive rate}, $\P\set{f(\r{x}) > t}$; in practice, all possible threshold settings are determined by the empirical distribution of $f(\r{x})$, so it is natural to use this distribution of scores to standardize the threshold scale. \citet{cv09a} derived consistency and asymptotic normality of an empirical PR curve estimate under certain regularity assumptions. However, in our work, we are exclusively focused on the development of finite-sample concentration bounds for predictive value curves. Previously, uniform bounds have been established that can be used to control collections of empirical conditional probabilities \cite{sbw09, bdfm19}; for fixed thresholds, predictive values can be seen as examples of such quantities, but the estimators of ppv and npv used in our paper employ \textit{data-dependent thresholds} computed using the full sample, and therefore they are not amenable to the same type of analysis done in these prior works. While it is possible to convert uniform bounds on tpr and tnr into bounds on ppv and npv if provided with an estimate of the label bias, the only such result for tpr and tnr that we found was given by \citet{cv10}, which bounds the sup norm $\lVert\widehat{\text{ROC}} - \text{ROC}\rVert_\infty$ for a particular estimator of the ROC curve. This is too loose for our purposes as it will yield a constant-width confidence band, whereas \textit{we seek bounds that are sensitive to, and adjust with, the positive rate.} 

In addition, the framework we have outlined also bears resemblance to the bipartite ranking problem in machine learning (\citet{mw16} provide a systematic theoretical study). Area under the ROC curve (AUC) is a popular way to summarize the performance of a scoring function in this context \cite{aghpr05, clv08, uag05}. There are several alternative objectives to the standard AUC that have been designed to place increased emphasis on specific regions of the ROC curve; some examples include the local AUC \cite{cv07}, the P-Norm Push \cite{r09}, and the Normalized Discounted Cumulative Gain, or NDCG \cite{wwlhcl13}. Other works study objectives that are closely related to the area under the PR curve \cite{hzhr02, yfrj07, s11, bep13, bcmr12}. 

The methods we employ to prove our results are similar to those used in the papers cited above to develop statistical guarantees for the bipartite ranking problem. Namely, we leverage McDiarmid's inequality \cite{m89} (included in Appendix A) in order to adapt symmetrization techniques from learning theory \cite{dgl96, bbl05, ab07, mp19} to our setting, which involves nonlinear predictive value statistics. However, aside from the technical similarities in proof methodology, there is an important conceptual distinction to be made. The persistent strategy in this line of ranking literature is to define a single-number summary for the empirical performance of scoring functions and develop theory for such a univariate statistic; in contrast, \textit{we analyze the full predictive value curves directly instead of attempting to collapse them into a single scalar of interest. This preserves the integrity and interpretability of ppv and npv, and the bounds we derive provide predictive value confidence intervals for all scoring functions, including those that can be obtained by optimizing the above objectives.}

\paragraph{Organization} The rest of the paper is organized as follows. In section 2, we describe the statistical model in detail and provide some theoretical motivation for restricting consideration to classifiers generated by level sets of scoring functions. In section 3, we prove a large deviation bound is used to construct distribution-free confidence bands for the predictive value curves of a fixed function $f$. In section 4, we prove a uniform convergence bound that extends the previous confidence band construction to the more general setting where the scoring function $f$ is selected from a class $\mathcal{F}$ in a data-dependent fashion. Section 5 provides some concluding remarks.

\section{Setup}

We consider the classical statistical learning model and review some standard notation. Let $\mathcal{X}$ be the instance space and $\mathcal{Y} = \set{0, 1}$ be the label space, and let $(\r{x}, \r{y}) \sim \P$ be an underlying joint distribution over $\mathcal{X}\times\mathcal{Y}$. Let $\mu$ be the marginal distribution over $\mathcal{X}$ and $\eta:\mathcal{X}\rightarrow [0, 1]$ be the regression function 
\[
\eta(x) = \P\set{\r{y} = 1 \given \r{x} = x}\;.
\]
The canonical goal in pattern recognition is to produce a data-dependent classifier $h:\mathcal{X}\rightarrow \mathcal{Y}$, $h(x) = \one{x\in A}$ for $A\subseteq \mathcal{X}$, with desirable generalization properties. The positive predictive value (ppv) and negative predictive value (npv) are two such properties
\begin{align*}
\ppv(A) &= \P\set{\r{y} = 1 \given \r{x} \in A} = \frac{\int_A \eta\, d\mu}{\int_A d\mu}\;,\\
\npv(A) &= \P\set{\r{y} = 0 \given \r{x} \not\in A} = \frac{\int_{A^c} (1 - \eta)\, d\mu}{\int_{A^c} d\mu}\;,
\end{align*}
which we take to equal zero if the conditioning events are zero-probability sets. These quantities are well-known in the statistics literature \cite{sg08, cv09}. Naturally, we are interested in sets with both high positive and high negative predictive values. At the population level, it can be shown that upper level sets of the function $\eta$ taking the form $A = \set{x: \eta(x) > t}$ are the only sets worth considering with regards to the predictive value criteria above. The classical Neyman-Pearson lemma \cite{mw16} establishes a similar result for different criteria, the Type I and Type II error rates. Before proving our claim, we introduce the following notation: 
\begin{itemize}
\item $A' \succeq A$ if $\ppv(A') \ge \ppv(A)$ and $\npv(A') \ge \npv(A)$,  
\item $A' \succ A$ if either $\ppv(A') \ge \ppv(A)$ and $\npv(A') > \npv(A)$, or $\ppv(A') > \ppv(A)$ and $\npv(A') \ge \npv(A)$.  
\end{itemize}

\textbf{Theorem 1} : \textit{ If $A = \set{x: \eta(x) > t}$ for some $t\in [0,1]$, then there is no set $A'$ such that $A' \succ A$. Furthermore, under the assumption that $\eta(\r{x})$ is a continuous random variable, for any set $A'$ there exists a set $A = \set{x: \eta(x) > t}$ with $\mu(A) = \mu(A')$ such that $A \succeq A'$.}

\textit{Proof. } See Appendix B. \qed

This theorem shows that the level sets of $\eta$ cannot be improved upon, and for any set $A'$ we can find a level set $A$ of $\eta$ which constitutes an improvement. In this sense, the level sets of $\eta$ (or any monotone increasing transformation of $\eta$) dominate all other subsets of the space $\mathcal{X}$, and, in particular, they dominate the level sets of any other function $f: \mathcal{X} \rightarrow \R$. If we assume that $f(\r{x})$ follows a continuous distribution, a natural way to re-parameterize the level sets of $f$ is via the quantile function defined as follows: for $\alpha \in [0,1]$
\[
q_f(\alpha) = \sup\set{t : \mu\set{x: f(x) > t} = \alpha}\;.
\] 
Then we can introduce variants of the $\ppv, \npv$ criteria adapted for use with functions, adapted from the definitions provided by \citet{mp04}:
\begin{align*}
\ppv(f, \alpha) &= \ppv(\set{x: f(x) > q_f(\alpha)})\;, \\[5px]
\npv(f, \alpha) &= \npv(\set{x: f(x) > q_f(\alpha)})\;.
\end{align*}
By construction, the right endpoint of the ppv curve is the label bias, $\ppv(f, 1) = \P\set{\r{y}=1}$, and the left endpoint of the npv curve is the complementary probability, $\npv(f, 0) = \P\set{\r{y}=0}$ . Furthermore, from Theorem 1, we know that $\ppv(\eta, \cdot) \ge \ppv(f, \cdot)$ and $\npv(\eta, \cdot) \ge \npv(f, \cdot)$ for all possible $f$. In the absence of oracle knowledge of $\eta$, this setup motivates the search for a function $f$ with high $\ppv(f, \alpha), \npv(f, \alpha)$ values across all settings of $\alpha \in [0,1]$. Thresholding the function at different $\alpha$ generates a nested sequence of subsets exhibiting a spectrum of predictive value behaviors, which allows the practitioner considerable flexibility in selecting a classifier that meets the needs of their specific problem. 

\paragraph{Example} For the purpose of illustration, we provide here a simple instantiation of the above setup -- this example will also be used to demonstrate results in the following section as well. Let $\mu = \mathcal{N}(0, I_d)$, $\eta(x) = \exp(-\norm{x}^2)$, and $f_1(x) = x[1]$ extracts the first coordinate of $x$. We can write down the predictive value curves of $\eta$ analytically:
\begin{align*}
\ppv(\eta, \alpha) &= \frac{1}{\alpha 3^{d/2}} \cdot \paren{1 - \frac{\Gamma(d/2\,,\, 3s_\alpha/2)}{\Gamma(d/2)}}\;, \\[5px]
\npv(\eta, \alpha) &= 1 - \frac{1}{(1-\alpha)3^{d/2}}\paren{\frac{\Gamma(d/2\,,\, 3s_\alpha/2)}{\Gamma(d/2)}}\;,
\end{align*}
where $s_\alpha = \log(1/q_\eta(\alpha))$ is the $\alpha$-percentile of the $\chi^2_d$ distribution, and $\Gamma(\cdot, \cdot)$ is the incomplete gamma function.  The predictive value curves for $f_1$ are
\begin{align*}
\ppv(f_1, \alpha) &= \frac{1}{\alpha 3^{(d-1)/2}} \paren{1 - \frac{1}{\sqrt{3}} \Phi(\sqrt{3}\, t_\alpha)}\;, \\[5px]
\npv(f_1, \alpha) &= 1 - \frac{1}{(1-\alpha)3^{(d-1)/2}} \paren{\frac{1}{\sqrt{3}} \Phi(\sqrt{3}\, t_\alpha)}\;,
\end{align*}
where $t_\alpha$ is the $(1-\alpha)$-percentile of the standard normal distribution Details of these calculations can be found in Appendix B.1. These curves are plotted in Figure \ref{fig:2}; we see that the predictive value curves of $\eta$ dominate the predictive value curves of $f_1$ as established in Theorem 1. 

\begin{figure}[t]
\centering
\includegraphics[scale = .4]{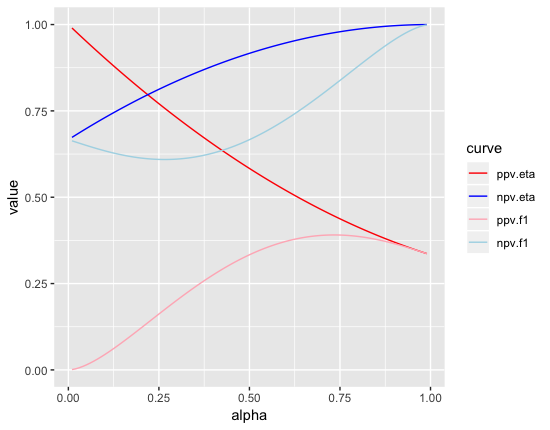}
\caption{Positive and negative predictive value curves for $\eta$ and $f_1$, with $d = 2$.}
\label{fig:2}
\end{figure}

\section{Large deviation bound}

The previous section provided theoretical motivation for the use of scoring functions to represent collections of subsets of the input space $\mathcal{X}$ in the predictive value framework. In this section, we address the following concern: given a fixed function $f:\mathcal{X}\rightarrow \R$, how accurately can we estimate its predictive value curves $\ppv(f, \cdot), \npv(f, \cdot)$ using an independent test data set? The type of analysis presented here takes inspiration from the techniques used by \citet{hzhr02} and \citet{aghpr05} to derive large deviation bounds for average precision and area under the ROC curve (AUC), respectively.      

Suppose we receive examples $(\r{x}_1, \r{y}_1), \ldots, (\r{x}_n, \r{y}_n)$ drawn i.i.d. from $\P$, and let $\alpha_k = k/n$ for $k = 1,2,\ldots, n-1$. Natural empirical estimates for the positive and negative predictive values are 
\begin{align*}
\widehat{\ppv}(f, \alpha_k) &= \frac{1}{k}\cdot \sum_{i=1}^n \one{\r{y}_i = 1\;, \; f(\r{x}_i) > \hat{q}_f(\alpha_k)} \\[5px]
\widehat{\npv}(f, \alpha_k) &= \frac{1}{n-k}\cdot \sum_{i=1}^n \one{\r{y}_i = 0\;, \; f(\r{x}_i) \le \hat{q}_f(\alpha_k)}
\end{align*}
where if $f_{(1)} \le f_{(2)} \le \cdots \le f_{(n)}$ is the list of values $f(\r{x}_1), f(\r{x}_2), \ldots, f(\r{x}_n)$ in sorted order, then the empirical quantiles are taken as 
\[
\hat{q}_f(\alpha_k) = \frac{f_{(n-k)} + f_{(n-k+1)}}{2} \qquad \text{for }\; k = 1,\ldots,n\;.
\]
An immediate question one might ask is: how far are these empirical quantities, in expectation, from their population counterparts? These quantities are ``almost" averages of iid Bernoulli variables aside from the dependence introduced by the empirical quantiles, and we will see that for relatively large values of $k$ they are ``almost" unbiased estimators. For the rest of this section, we will assume that $f(\r{x})$ follows a continuous distribution, so that the quantile function $q_f(\cdot)$ is well-defined.

\textbf{Lemma 1} : \textit{For any $k = 1,2, \ldots, n-1$ we have that}
\begin{align*}
\abs{\E\brack{\widehat{\ppv}(f, \alpha_k)} - \ppv(f, \alpha_k)}&\le \frac{n}{2k} \sqrt{\frac{\pi}{2(n-1)}}\;,\\[5px]
\abs{\E\brack{\widehat{\npv}(f, \alpha_k)} - \npv(f, \alpha_k)}&\le \frac{n}{2(n-k)}\sqrt{\frac{\pi}{2(n-1)}}\;.
\end{align*}

\textit{Proof. } See Appendix C. \qed

The key step of the proof involves an application of Hoeffding's inequality (included in Appendix A). Now that we have established control over the bias of the empirical predictive values, our next task is bounding the variance of these estimates. This is easily done by applying McDiarmid's inequality (included in Appendix A), as shown in the following theorem, which contains our distribution-free large deviation concentration inequality for predictive values.  

\textbf{Theorem 2} : \textit{With probability at least $1 - \delta$, for all $k = 1,2, \ldots, n-1$}
\begin{align*}
\abs{\widehat{\ppv}(f, \alpha_k) - \ppv(f, \alpha_k)} &\le \frac{1}{k}\sqrt{\frac{n\log(4n/\delta)}{2}} + \frac{n}{2k}\sqrt{\frac{\pi}{2(n-1)}} \\[5px]
&\le O\paren{\frac{1}{\alpha_k}\sqrt{\frac{\log(n) + \log(1/\delta)}{n}}}\;,\\[5px]
\abs{\widehat{\npv}(f, \alpha_k) - \npv(f, \alpha_k)} &\le \frac{1}{n-k}\sqrt{\frac{n\log(4n/\delta)}{2}} + \frac{n}{2(n-k)}\sqrt{\frac{\pi}{2(n-1)}} \\[5px]
&\le O\paren{\frac{1}{1-\alpha_k}\sqrt{\frac{\log(n) + \log(1/\delta)}{n}}}\;.
\end{align*}

\textit{Proof. } It can be seen that perturbing any data pair $(x_i, y_i)$ changes the value of $\widehat{\ppv}(f, \alpha_k)$ by at most $1/k$ and $\widehat{\npv}(f, \alpha_k)$ by at most $1/(n-k)$, so McDiarmid's inequality implies that
\begin{align*}
\P\set{\abs{\widehat{\ppv}(f, \alpha_k) - \E\brack{\widehat{\ppv}(f, \alpha_k)}} > \epsilon} &\le 2\cdot \exp\paren{-\frac{2\epsilon^2k^2}{n}}\;, \\[5px]
\P\set{\abs{\widehat{\npv}(f, \alpha_k) - \E\brack{\widehat{\npv}(f, \alpha_k)}} > \epsilon} &\le 2\cdot \exp\paren{-\frac{2\epsilon^2(n-k)^2}{n}}\;.
\end{align*}
Passing to a confidence interval representation by setting these bounds equal to $\delta/2n$, we obtain the desired simultaneous guarantee after applying the union bound together with Lemma 1. \qed

From Theorem 2, we see that the confidence interval for $\ppv(f, \alpha_k)$ tightens as $\alpha_k$ increases, and similarly, the confidence interval for $\npv(f, \alpha_k)$ tightens as $\alpha_k$ decreases. Intuitively, this makes sense; there is very little data available to estimate ppv for small values of $\alpha_k$ but plentiful data for large values of $\alpha_k$, and vice versa with regards to npv. Furthermore, modulo log factors, we can only obtain nontrivial guarantees simultaneously for both ppv and npv when $\sqrt{n} \ll k \ll n - \sqrt{n}$, or equivalently, when $ 1/\sqrt{n} \ll \alpha_k \ll 1 - 1/\sqrt{n}$. In fact, we need $\alpha_k$ to be strictly bounded away from 0 and 1 in order to achieve a $1/\sqrt{n}$ rate of convergence for the empirical predictive values. Any rate of decay with $n$ in either direction, $\alpha_k = o(1)$ or $1 - \alpha_k = o(1)$, will result in a degradation of the $1/\sqrt{n}$ rate. We state this explicitly as a corollary to the theorem.

\textbf{Corollary 1} : \textit{Let $c$ be a constant, independent of $n$, satisfying $0 < c < 1$. With probability at least $1-\delta$, for all $k$ such that $c n \le k \le (1-c) n$}
\begin{align*}
\abs{\widehat{\ppv}(f, \alpha_k) - \ppv(f, \alpha_k)} &\le O\paren{\frac{1}{c}\sqrt{\frac{\log(n) + \log(1/\delta)}{n}}}\;,\\[5px]
\abs{\widehat{\npv}(f, \alpha_k) - \npv(f, \alpha_k)} &\le O\paren{\frac{1}{c}\sqrt{\frac{\log(n) + \log(1/\delta)}{n}}}\;.
\end{align*}

\textit{Proof. } Follows directly from Theorem 2. \qed

\begin{figure}[b]
\centering
\includegraphics[scale = .35]{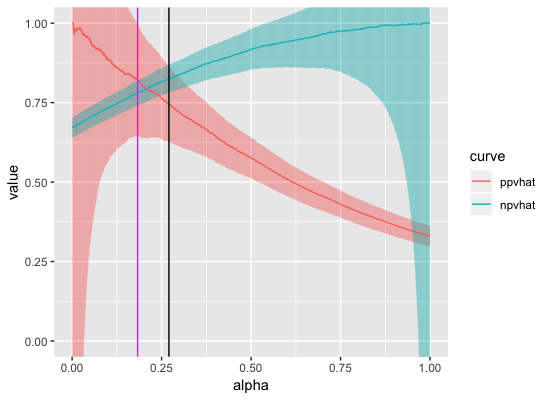}
\includegraphics[scale = .35]{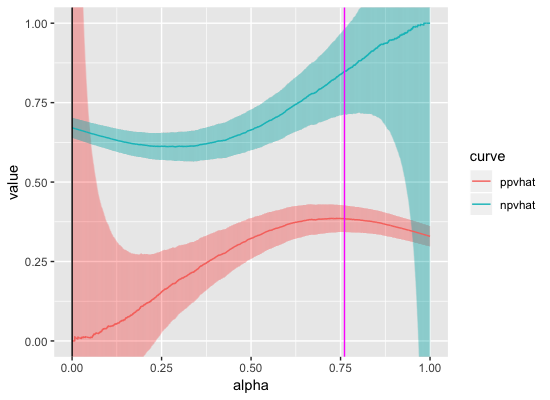}

\caption{Empirical predictive value curves for $\eta$ (left) and $f_1$ (right), with $n = 10^4, d = 2, \delta = .05$}. 
\label{fig:3}
\end{figure}

To see the bound in action, we simulate a data set of size $n = 10,000$ from the distribution in Example 2.1; in Figure \ref{fig:3} we plot the $\widehat{\ppv}, \widehat{\npv}$ empirical curves for both $\eta(x) = \exp(-\norm{x}^2)$ and $f_1(x) = x[1]$, along with error bars implied by the inequalities of Theorem 2.

These results advocate exercising caution when selecting a threshold level; if one selects too small or too large an $\alpha_k$, one runs the risk of losing statistical control of the predictive values. We note that it is standard practice in many machine learning applications  to choose the threshold $\alpha^*$ that produces a level set of $f$ with maximum classification accuracy:
\[
\alpha^* = \argmax_{\alpha_k} \;\; \alpha_k\cdot \widehat{\ppv}(f, \alpha_k) + (1 - \alpha_k)\cdot \widehat{\npv}(f, \alpha_k)\;.
\]

This threshold setting does not take into account the uncertainty in the empirical predictive values, and the objective formulation assumes that the relative importance of positive predictive values coincides with the positive rate $\alpha_k$. A more agnostic and statistically sound selection would be 
\[
\alpha^\star = \argmax_{\alpha_k} \;\; \min\set{\widehat{\text{ppvLCB}}(f,\alpha_k)\; , \;\widehat{\text{npvLCB}}(f,\alpha_k)}\;,
\]
where $\widehat{\text{ppvLCB}}(f,\alpha_k)$ and $\widehat{\text{npvLCB}}(f,\alpha_k)$ denote the lower confidence bounds for $\ppv(f, \alpha_k)$ and $\npv(f, \alpha_k)$, respectively, provided by Theorem 2. In Figure \ref{fig:3}, we include a vertical black line for $\alpha^*$ and a vertical pink line for $\alpha^\star$. The difference between the two thresholds is quite stark for the scoring function $f_1(x) = x[1]$; the setting $\alpha^* = 1/n$ fails to provide any useful information about ppv, whereas $\alpha^\star$ falls within the range to produce nontrivial confidence intervals for both ppv and npv. This simulation example illustrates how the confidence bands developed in this section can enable and motivate principled methods of threshold selection that take into account statistical significance of the empirical positive and negative predictive values. 


\section{Uniform convergence bound}

While the previous results are useful for evaluating a fixed function $f$ on an independent sequence of test data, often times in machine learning problems, one would prefer to use most of the available data to select the function $f$ instead of reserving a significant chunk solely for validation. The standard approach to establishing generalization performance guarantees from the same data used to select $f$ is via uniform convergence bounds over a function class $\mathcal{F}$ \cite{bbl05, aghpr05}. In this section, we will prove a distribution-free bound for predictive values that holds uniformly over the choice of scoring function $f\in \mathcal{F}$ and threshold level $\alpha \in [0,1]$. This kind of result holds independently of any algorithm used by the learner, and thus it can account for any arbitrarily complicated process implemented to make data-dependent selections of $f$ and $\alpha$. 

We will need some additional notation before we proceed. Let $\mathcal{S} = \mathcal{X}\times \set{0, 1}$ denote the sample space, let $(x, y)$ represent a data sequence $((x_1, y_1), (x_2, y_2), \ldots, (x_n, y_n)) \in \mathcal{S}^n$, and for $f:\mathcal{X}\rightarrow \R$, $f(x)$ denotes the vector $(f(x_1), f(x_2), \ldots, f(x_n))$. The map $\phi_k: \R^n \rightarrow 2^{[n]}$ takes a real vector to the subset of indices that correspond to the largest $k$ values in the vector, where ties are broken using the original order (a so-called stable sort); for example, $\phi_2(-1.3, 2, -5, 1.2) = \set{2, 3}$. In addition, we define the $(n,k)$-\textit{order coefficient} to be the maximum number of such size-$k$ subsets realized by the function class:
\[
\Theta(\mathcal{F}, n, k) = \max_{x\in \mathcal{X}^n} \, |\set{\phi_k(f(x)): f\in \mathcal{F}} |\;.
\]
We will need a sub-exponential bound on this coefficient in order for our uniform convergence result to provide a meaningful confidence band. One way to obtain such a bound is to simply note that $\Theta(\mathcal{F}, n, \alpha_k n)\le \Pi(\mathcal{F}, n)$, where $\Pi(\mathcal{F}, n)$ denotes the shattering number
\[
\Pi(\mathcal{F}, n) = \max_{x\in \mathcal{X}^n} \abs{\set{\text{sign}(f(x) - b): f\in \mathcal{F}, \, b\in \R^n}}
\]
and $\text{sign}(v)$ returns the binary vector $(\one{v_1 > 0}, \ldots, \one{v_n > 0})$.  For classes $\mathcal{F}$ with finite VC-subgraph dimension
\[
d = \text{VC}(\mathcal{F}) = \max\set{n : \Pi(\mathcal{F}, n) = 2^n}\;,
\]
the Sauer-Shelah lemma states that 
\[
\Pi(\mathcal{F}, n) \le \sum_{i=0}^{d} \binom{n}{i} \le \paren{\frac{e n}{d}}^d\;,
\]
which yields a polynomial upper bound for the order coefficient. Derivations of this classical lemma can be found in many standard references \cite{vw96, gkkw02}. Now, we are ready to state the main theorem of this work. 

\textbf{Theorem 3 : } With probability at least $1 - \delta$, for $k = 1, 2, \ldots, n-1$ and for all $f\in \mathcal{F}$ with $d = \text{VC}(\mathcal{F}) < \infty$
\begin{align*}
\abs{\widehat{\ppv}(f, \alpha_k) - \ppv(f, \alpha_k)} &\le \frac{1}{k}\sqrt{2n\cdot \log(8n \cdot \Theta(\mathcal{F}, n, k)^2/\delta)} + \frac{n}{2k} \sqrt{\frac{\pi}{2(n-1)}} \\[5px]
&\le O\paren{\frac{1}{\alpha_k}\sqrt{\frac{d\cdot \log(n/d) + \log(1/\delta)}{n}}}\;, \\[5px]
\abs{\widehat{\npv}(f, \alpha_k) - \npv(f, \alpha_k)} &\le \frac{1}{n-k}\sqrt{2n\cdot \log(8n \cdot \Theta(\mathcal{F}, n, n-k)^2 / \delta)} + \frac{n}{2(n-k)} \sqrt{\frac{\pi}{2(n-1)}}\\[5px]
&\le O\paren{\frac{1}{1-\alpha_k}\sqrt{\frac{d\cdot \log(n/d) + \log(1/\delta)}{n}}}\;.
\end{align*}

\textit{Proof. } See Appendix D. \qed

The method of proof utilizes modified versions of classical symmetrization techniques and makes critical use of McDiarmid's inequality (included in Appendix A) to control the nonlinear empirical predictive value statistics. We see the same form of threshold dependence as in Theorem 2: the ppv confidence band tightens as $\alpha_k$ increases, and the npv confidence band tightens as $\alpha_k$ decreases. Up to log factors, the bound implies consistency of the predictive value curves uniformly across the function class $\mathcal{F}$ in the regime where $\sqrt{d/n} \ll \alpha_k \ll 1 - \sqrt{d/n}$. We also note, as we did in the previous section, if $\alpha_k = o(1)$ or $1 - \alpha_k = o(1)$ as $n/d$ grows, then the $\sqrt{d/n}$ rate of convergence will suffer. 

\textbf{Corollary 2} : \textit{Let $c$ be a constant, independent of $n$, satisfying $0 < c < 1$. With probability at least $1-\delta$, for all $k$ such that $c n \le k \le (1-c) n$}
\begin{align*}
\abs{\widehat{\ppv}(f, \alpha_k) - \ppv(f, \alpha_k)} &\le O\paren{\frac{1}{c}\sqrt{\frac{d\cdot \log(n/d) + \log(1/\delta)}{n}}}\;, \\[5px]
\abs{\widehat{\npv}(f, \alpha_k) - \npv(f, \alpha_k)} &\le O\paren{\frac{1}{c}\sqrt{\frac{d\cdot \log(n/d) + \log(1/\delta)}{n}}}\;.
\end{align*}

\textit{Proof. } Follows directly from Theorem 3. \qed

The VC-subgraph dimension is a general property that can be controlled in many interesting cases. The following proposition, which is a well-known result, bounds the VC-subgraph dimension for finite-dimensional vector spaces. 

\textbf{Proposition 1 (Lemma 2.6.15 in \cite{vw96})} : \textit{Let $\mathcal{F}$ be any finite-dimensional vector space of functions. Then }
\begin{align*}
\text{VC}(\mathcal{F}) \le \text{dim}(\mathcal{F}) + 1\;.
\end{align*}

This bound can be used in conjunction with Theorem 3 to construct predictive value confidence bands for a variety of practical classes of scoring functions; for example, linear half-spaces and ellipsoids can be expressed as level sets of functions from finite-dimensional vector spaces \cite{w19}. In addition, Lemma 2.6.18 \cite{vw96} ensures that the VC-subgraph property is preserved under simple arithmetic operations for combining classes of functions. These lemmas generate an ecosystem of VC-subgraph classes for which the uniform convergence bounds in this section provide nontrivial statistical guarantees.

\section{Conclusion}

In this paper, we studied predictive value performance of scoring functions from the viewpoint of statistical learning theory. We showed that the level-set approach to classification is theoretically well-motivated with regards to positive and negative predictive values. The large deviation bound established in Section 3 is used to construct a confidence band around the ppv and npv curves of a given scoring function $f$; this result enables the learner to threshold $f$ in a principled manner by using fresh data to select a level $\alpha$ that balances the magnitudes of ppv and npv with the widths of their respective confidence intervals. Our Theorem 2 implies consistency of the empirical predictive value curves in the regime where the threshold $\alpha$ is not too small or large; namely, we require $1/\sqrt{n} \ll \alpha \ll 1 - 1/\sqrt{n}$. Finally, when the scoring function itself is chosen in a data-dependent fashion, we need a more general confidence band that holds uniformly over a function class $\mathcal{F}$. The main result of our work, Theorem 3, derives the desired uniform confidence band, and establishes consistency of the predictive value curves for function classes with finite VC-subgraph dimension $d$ when the threshold level satisfies $\sqrt{d/n} \ll \alpha \ll 1-\sqrt{d/n}$. Unlike existing theory that has been developed in the bipartite ranking literature, our results address concentration of the entire predictive value curves instead of specific scalar functionals of these curves. This level of generality enables our bounds to provide confidence intervals for the predictive values of classifiers and scoring functions obtained via any algorithm.


%
%

{
\small
\bibliography{mybib}{}

}

\vspace{1cm}

\section*{A. Useful inequalities}

\textbf{Theorem (Hoeffding's Inequality)} : \textit{Let $X_1, \ldots, X_n$ be independent random variables, each of which is strictly bounded in an interval $a_i \le X_i \le b_i$, and let $S_n = \sum_{i=1}^n X_i$ denote their sum. Then we have} 
\[
\P\set{S_n - \E\brack{S_n} \ge \epsilon} \le \exp\paren{-\frac{2\epsilon^2}{\sum_{i=1}^n (b_i - a_i)^2}}\;.
\]

\textbf{Theorem (McDiarmid's Inequality)} : \textit{Let $X_1, \ldots, X_n$ be independent random variables taking values in some space $\mathcal{X}$. Let $g:\mathcal{X}^n \rightarrow \R$ be a function satisfying the bounded differences property: for all indices $i\in [n]$, all tuples $(x_1, \ldots, x_n)$, and all $x_i'\in \mathcal{X}$}
\[
\abs{g(x_1, \ldots, x_{i-1}, x_i, x_{i+1}, \ldots, x_n) - g(x_1, \ldots, x_{i-1}, x_i', x_{i+1}, \ldots, x_n)} \le c_i\;.
\]
\textit{Then we have }
\[
\P\set{g(X_1, \ldots, X_n) - \E\brack{g(X_1, \ldots, X_n)} \ge \epsilon} \le \exp\paren{-\frac{2\epsilon^2}{\sum_{i=1}^n c_i^2}}\;.
\]

\section*{B. Proof of Theorem 1}

\textbf{Theorem 1} : \textit{ If $A = \set{x: \eta(x) > t}$ for some $t\in [0,1]$, then there is no set $A'$ such that $A' \succ A$. Furthermore, under the assumption that $\eta(\r{x})$ is a continuous random variable, for any set $A'$ there exists a set $A = \set{x: \eta(x) > t}$ with $\mu(A) = \mu(A')$ such that $A \succeq A'$.}

\textit{Proof. } If $A' \succ A$ then we need either $\ppv(A) < \ppv(A')$ or $\npv(A) < \npv(A')$. Plugging in the definitions of the quantities involved, the first option translates into
\begin{align*}
\frac{\int_A \eta\, d\mu}{\int_A d\mu}  &< \frac{\int_{A'} \eta\, d\mu}{\int_{A'} d\mu} 
= \frac{\int_{A} \eta\, d\mu + \int_{A'\setminus A} \eta\, d\mu - \int_{A \setminus A'} \eta\, d\mu}{\int_{A} d\mu + \int_{A'\setminus A} d\mu - \int_{A \setminus A'} d\mu}\;,
\end{align*}
which, after some simplification, yields the inequality
\begin{align*}
\int_A \eta\, d\mu \cdot \paren{\int_{A'\setminus A} d\mu - \int_{A \setminus A'} d\mu} < \int_A d\mu \cdot \paren{\int_{A'\setminus A} \eta\, d\mu - \int_{A \setminus A'} \eta\, d\mu}\;.
\end{align*}
Recalling the definition of $A$, we arrive at the implication
\begin{equation}
\int_A \eta\, d\mu \cdot \paren{\mu(A') - \mu(A)} < t\cdot \mu(A) \paren{\mu(A') - \mu(A)}\;.
\end{equation}
Now we repeat these steps for the second option ($\npv(A) < \npv(A')$). Plugging in the definitions as before: 
\begin{align*}
\frac{\int_{A^c} (1 - \eta)\, d\mu}{\int_{A^c} d\mu}  &< \frac{\int_{A'^c} (1-\eta)\, d\mu}{\int_{A'^c} d\mu} 
= \frac{\int_{A^c} (1-\eta)\, d\mu + \int_{A'^c \setminus A^c} (1-\eta)\, d\mu - \int_{A^c \setminus A'^c} (1-\eta)\, d\mu}{\int_{A^c} d\mu + \int_{A'^c \setminus A^c} d\mu - \int_{A^c \setminus A'^c} d\mu}
\end{align*}
and upon simplifying we see that 
\begin{align*}
&\int_{A^c} (1 - \eta)\, d\mu \cdot \paren{\int_{A'^c \setminus A^c} d\mu - \int_{A^c \setminus A'^c} d\mu}  \\[5px]
& \qquad \qquad \qquad < \int_{A^c} d\mu \cdot \paren{\int_{A'^c \setminus A^c} (1-\eta)\, d\mu - \int_{A^c \setminus A'^c} (1-\eta)\, d\mu}\;.
\end{align*}
Once again recalling the definition of $A$ and further simplification yields:
\begin{equation}
\int_{A^c} (1 - \eta)\, d\mu \cdot \paren{\mu(A) - \mu(A')} < (1-t)\cdot \mu(A^c) \paren{\mu(A) - \mu(A')}\;.
\end{equation}
Now we can argue by casework on $\mu(A)$ and $\mu(A')$. If $\mu(A) = \mu(A')$, both (1) and (2) imply $0 < 0$ which is immediately contradictory, so we can assume $\mu(A) \ne \mu(A')$. If $\mu(A) > \mu(A')$ then we can cancel the factor of $\paren{\mu(A) - \mu(A')}$ from both sides of (2) and lower bound the left hand side by $(1-t)\cdot \mu(A^c)$ to obtain a contradiction. Similarly, if $\mu(A) < \mu(A')$ then we can cancel the factor $\paren{\mu(A) - \mu(A')}$ from both sides of (1) and lower bound the left hand side by $t\cdot \mu(A)$ to obtain a contradiction. This completes the first part of the proof.  

For the second part, since we assume $\eta(\r{x})$ is a continuous random variable, the function $g(t) = \mu\set{x: \eta(x) > t}$ is continuous in $t$ with $g(0) = 1$ and $g(1) = 0$. This implies the existence of a $t^*\in [0,1]$ such that $g(t^*) = \mu(A')$, so we can take $A = \set{x: \eta(x) > t^*}$. By construction, we have that $\mu(A) = \mu(A')$. If $\mu(A) = 0$ or $\mu(A) = 1$ then we immediately have $\ppv(A) = \ppv(A')$ and $\npv(A) = \npv(A')$, so assume that $\mu(A) \not\in\set{0, 1}$. We now have 
\begin{align*}
\ppv(A) - \ppv(A') &= \frac{\int_{A}\eta \, d\mu}{\int_{A}d\mu} - \frac{\int_{A'} \eta\, d\mu}{\int_{A'} d\mu} \\[5px]
&= \frac{1}{\mu(A)}\cdot \paren{\int_{A\setminus A'}\eta \, d\mu - \int_{A'\setminus A} \eta\, d\mu} \\[5px]
&\ge \frac{1}{\mu(A)}\cdot t^*(\mu(A) - \mu(A')) = 0\;,
\end{align*}

where the last line utilized the definition of $A$. Similarly,

\begin{align*}
\npv(A) - \npv(A') &= \frac{\int_{A^c} (1 - \eta)\, d\mu}{\int_{A^c}d\mu} - \frac{\int_{A'^c}(1 - \eta)\, d\mu}{\int_{A'^c}d\mu}\\[5px]
&= \frac{1}{\mu(A^c)} \cdot \paren{\int_{A'^c\setminus A^c} \eta\, d\mu - \int_{A^c\setminus A}\eta\, d\mu} \\[5px]
&\ge \frac{1}{\mu(A^c)}\cdot t^*(\mu(A'^c) - \mu(A^c)) = 0 \;.
\end{align*}

We conclude that $A\succeq A'$, and the proof is complete. 
\qed

\subsection*{B.1 Derivation of example}

We can explicitly calculate the $\ppv$ curve of $\eta$ as follows:
\begin{align*}
\ppv(\eta, \alpha) &= \P\set{\r{y} = 1 \given \eta(\r{x}) > q_\eta(\alpha)} \\
&= \frac{1}{\alpha}\int \eta(x)\cdot \one{\eta(x) > q_\eta(\alpha)} \d{\mu(x)}\\
&= \frac{1}{\alpha}\int \exp(-\norm{x}^2)\cdot \one{\norm{x}^2 \le  \log(1/q_\eta(\alpha))}\d{\mathcal{N}(x; 0, I_d)} \\
&= \frac{1}{\alpha} \int_0^{s_\alpha} \exp(-r) \d{\chi^2_d(r)} \\[5px]
&= \frac{1}{\alpha 3^{d/2}} \cdot \paren{1 - \frac{\Gamma(d/2\,,\, 3s_\alpha/2)}{\Gamma(d/2)}}
\end{align*}
where $s_\alpha = \log(1/q_\eta(\alpha))$ is the $\alpha$-percentile of the $\chi^2_d$ distribution, and $\Gamma(\cdot, \cdot)$ is the incomplete gamma function. Similarly, the $\npv$ curve of $\eta$ follows suit:
\begin{align*}
\npv(\eta, \alpha) &= \P\set{\r{y} = 0 \given \eta(\r{x}) \le q_\eta(\alpha)}\\[5px]
&= \frac{1}{1-\alpha} \int (1 - \eta(x)) \cdot \one{\eta(x) \le q_\eta(\alpha)} \d{\mu(x)} \\[5px]
&= 1 - \frac{1}{1-\alpha}\int_{s_\alpha}^\infty \exp(-r) \d{\chi^2_d(r)} \\[5px]
&= 1 - \frac{1}{(1-\alpha)3^{d/2}}\paren{\frac{\Gamma(d/2\,,\, 3s_\alpha/2)}{\Gamma(d/2)}}\;.
\end{align*}

The ppv curve for $f$ is
\begin{align*}
\ppv(f, \alpha) &= \frac{1}{\alpha}\int \exp(-\norm{x}^2)\cdot \one{x[1] > q_f(\alpha)} \d{\mu(x)} \\[5px]
&= \frac{1}{\alpha}\int \exp(-(x[1]^2 + \norm{x[-1]}^2))\cdot \one{x[1] > q_f(\alpha)} \d{\mu(x)} \\[5px]
&= \frac{1}{\alpha 3^{(d-1)/2}} \int \exp(-x[1]^2) \cdot \one{x[1] > q_f(\alpha)} \d{\mu(x[1])} \\[5px]
&= \frac{1}{\alpha 3^{(d-1)/2}} \paren{1 - \frac{1}{\sqrt{3}} \Phi(\sqrt{3}\, t_\alpha)} 
\end{align*}
where $x[-1]$ denotes all coordinates except $x[1]$, and $t_\alpha$ is the $(1-\alpha)$-percentile of the standard normal distribution. Similarly, the npv curve for $f$ is 
\begin{align*}
\npv(f, \alpha) &= 1 - \frac{1}{1-\alpha}\int \exp(-\norm{x}^2)\cdot \one{x[1] \le q_f(\alpha)} \d{\mu(x)} \\[5px]
&= 1 - \frac{1}{1-\alpha}\int \exp(-(x[1]^2 + \norm{x[-1]}^2))\cdot \one{x[1] \le q_f(\alpha)} \d{\mu(x)} \\[5px]
&= 1 - \frac{1}{(1-\alpha)3^{(d-1)/2}} \paren{\frac{1}{\sqrt{3}} \Phi(\sqrt{3}\, t_\alpha)}\;.
\end{align*}

\section*{C. Proof of Lemma 1}
\label{app:B}

\textbf{Lemma 1} : \textit{For any $k = 1,2, \ldots, n-1$ we have that}
\begin{align*}
\abs{\E\brack{\widehat{\ppv}(f, \alpha_k)} - \ppv(f, \alpha_k)}&\le \frac{n}{2k} \sqrt{\frac{\pi}{2(n-1)}}\;,\\[5px]
\abs{\E\brack{\widehat{\npv}(f, \alpha_k)} - \npv(f, \alpha_k)}&\le \frac{n}{2(n-k)} \sqrt{\frac{\pi}{2(n-1)}}\;.
\end{align*}

\textit{Proof. } We will prove the first inequality. The second bound follows from a completely analogous set of arguments. First, we observe that by linearity of expectation, 
\[
\E\brack{\widehat{\ppv}(f, \alpha_k)} = \frac{n}{k}\cdot \P^n\set{\r{y}_1 = 1\;, \; f(\r{x}_1) > \hat{q}_f(\alpha_k)}
\]
and we also have that
\[
\ppv(f, \alpha_k) = \P\set{\r{y} = 1 \given f(\r{x}) > q_f(\alpha_k)} = \frac{n}{k}\cdot \P\set{\r{y} = 1\;,\; f(\r{x}) > q_f(\alpha_k)}
\]
so it suffices to establish the bound up to the common factor of $n/k$. We proceed with the following calculation:
\begin{align*}
&\abs{\P^n\set{\r{y}_1 = 1\;, \; f(\r{x}_1) > \hat{q}_f(\alpha_k)} - \P\set{\r{y} = 1\;,\; f(\r{x}) > q_f(\alpha_k)}} \\[5px]
&\qquad = \abs{\int \one{y_1 = 1\;, \; f(x_1) > \hat{q}_f(\alpha_k)}\, \d{\P^n} - \int \one{y_1 = 1\;,\; f(x_1) > q_f(\alpha_k)} \d{\P}} \\[5px]
&\qquad \le \int \abs{\one{f(x_1) > \hat{q}_f(\alpha_k)} - \one{f(x_1) > q_f(\alpha_k)}} \d{\P^n } \\[5px]
&\qquad \le \P^n\set{\hat{q}_f(\alpha_k) < f(\r{x}_1) \le q_f(\alpha_k)} + \P^n\set{q_f(\alpha_k) < f(\r{x}_1) \le \hat{q}_f(\alpha_k)} \\[5px]
&\qquad = \E\brack{\P^n\set{\hat{q}_f(\alpha_k) < f(\r{x}_1) \le q_f(\alpha_k) \given \r{x}_1}} + \E\brack{\P^n\set{q_f(\alpha_k) < f(\r{x}_1) \le \hat{q}_f(\alpha_k) \given \r{x}_1}} \\[5px]
&\qquad \le \E\brack{\P^n\set{\sum_{i = 2}^n \one{f(\r{x}_i) \ge f(\r{x}_1)} \le k-1 \given \r{x}_1} \cdot \one{f(\r{x}_1) \le q_f(\alpha_k)}} \\[5px]
&\qquad \qquad + \E\brack{\P^n\set{\sum_{i=2}^n \one{f(\r{x}_i) \ge f(\r{x}_1)} \ge k \given \r{x}_1} \cdot \one{f(\r{x}_1) > q_f(\alpha_k)}} \\[5px]
&\qquad =  \E\brack{\P^n\set{\sum_{i = 2}^n \one{f(\r{x}_i) \ge f(\r{x}_1)} - (n-1)(\alpha_k + \r{\delta}_0) \le k-1 - (n-1)(\alpha_k + \r{\delta}_0) \given \r{x}_1} \right. \\
&\left. \vphantom{\sum_{i=1}^n}\hspace{9cm} \times \one{f(\r{x}_1) \le q_f(\alpha_k)}} \\[5px]
&\qquad \qquad + \E\brack{\P^n\set{\sum_{i=2}^n \one{f(\r{x}_i) \ge f(\r{x}_1)} - (n-1)(\alpha_k - \r{\delta}_1) \ge k - (n-1)(\alpha_k - \r{\delta}_1) \given \r{x}_1} \right. \\
&\left. \vphantom{\sum_{i=1}^n} \hspace{9cm} \times \one{f(\r{x}_1) > q_f(\alpha_k)}} \\[5px]
&\qquad \le  \E\brack{ \exp\paren{-\frac{2}{n-1}\paren{\frac{k}{n} - (n-1)\r{\delta}_0 - 1}^2}\cdot \one{f(\r{x}_1) \le q_f(\alpha_k)}} \\[5px]
&\qquad \qquad + \E\brack{ \exp\paren{-\frac{2}{n-1}\paren{\frac{k}{n} + (n-1)\r{\delta}_1}^2}\cdot \one{f(\r{x}_1) > q_f(\alpha_k)}} 
\end{align*}

where $\r{\delta}_0 = \mu\set{x: f(\r{x}_1) \le f(x) \le q_f(\alpha_k)}$ and $\r{\delta}_1 = \mu\set{x: q_f(\alpha_k) \le f(x) \le f(\r{x}_1)}$, and the last line follows from an application of Hoeffding's inequality. Upon inspection, we see that $\r{\delta}_0, \r{\delta}_1$ follow the conditional distributions
\begin{align*}
\r{\delta}_0 \mid \set{f(\r{x}_1) \le q_f(\alpha_k)} & \sim \text{unif}[0\,,\, 1]\;, \\[5px]
\r{\delta}_1 \mid \set{f(\r{x}_1) > q_f(\alpha_k)} &\sim \text{unif}[0 \,,\, 1]\;,
\end{align*}
which allows us to write the last line of our calculation as
\begin{align*}
&\paren{1 - \frac{k}{n}}\cdot \int_0^1 \exp\paren{-\frac{2}{n-1}\paren{1 - \frac{k}{n} + (n-1)u}^2} \d{u} \\
&\hspace{6cm} + \frac{k}{n} \cdot \int_0^1 \exp\paren{-\frac{2}{n-1}\paren{\frac{k}{n} + (n-1)u}^2} \d{u} \\[5px]
&\qquad = \paren{1 - \frac{k}{n}}\cdot \frac{1}{\sqrt{2(n-1)}} \int_{\sqrt{\frac{2}{n-1}}(1 - k/n)}^{\sqrt{\frac{2}{n-1}}(n-k/n)} \exp(-v^2)\d{v} \\
&\hspace{6cm} + \frac{k}{n}\cdot \frac{1}{\sqrt{2(n-1)}} \int_{\sqrt{\frac{2}{n-1}}(k/n)}^{\sqrt{\frac{2}{n-1}}(n - 1 + k/n)} \exp(-v^2)\d{v} \\
&\qquad \le \frac{1}{\sqrt{2(n-1)}}\cdot \frac{\sqrt{\pi}}{2}\;.
\end{align*}
Combining this result with the factor of $n/k$ from our initial observation completes the proof. \qed

\section*{D. Proof of Theorem 3}

\textbf{Theorem 3 : } With probability at least $1 - \delta$, for $k = 1, 2, \ldots, n$ and for all $f\in \mathcal{F}$
\begin{align*}
\abs{\widehat{\ppv}(f, \alpha_k) - \ppv(f, \alpha_k)} &\le \frac{1}{k}\sqrt{2n\cdot \log(8n \cdot \Theta(\mathcal{F}, n, k)^2/\delta)} + \frac{n}{2k} \sqrt{\frac{\pi}{2(n-1)}} \\[5px]
\abs{\widehat{\npv}(f, \alpha_k) - \npv(f, \alpha_k)} &\le \frac{1}{n-k}\sqrt{2n\cdot \log(8n \cdot \Theta(\mathcal{F}, n, n-k)^2 / \delta)} + \frac{n}{2(n-k)} \sqrt{\frac{\pi}{2(n-1)}}\;. 
\end{align*}

\textit{Proof. } Let $\mathcal{S} = \mathcal{X}\times \set{0, 1}$ denote the sample space, let $(x, y)$ represent a data sequence $((x_1, y_1), (x_2, y_2), \ldots, (x_n, y_n)) \in \mathcal{S}^n$, and for $f:\mathcal{X}\rightarrow \R$, $\phi_k(f(x))$ denotes the vector $(f(x_1), f(x_2), \ldots, f(x_n))$. The map $\phi_k: \R^n \rightarrow 2^{[n]}$ takes a real vector to the subset of indices that correspond to the largest $k$ values in the vector, where ties are broken using the original order (a so-called stable sort); for example, $\phi_2(-1.3, 2, -5, 1.2) = \set{2, 3}$. For any collection of indices $C\in 2^{[n]}$, we define
\[
\bar{y}(C) = \frac{1}{|C|}\cdot  \sum_{i \in C} y_{i}
\]
which, when $C = \phi_k(f(x))$, is a more general and streamlined notation for $\widehat{\ppv}(f, \alpha_k)$. Now we want to prove a bound on the probability of the event
\[
A_{k,\epsilon} = \set{(x, y)\in \mathcal{S}^n: \exists f\in \mathcal{F} \st \abs{\bar{y}(\phi_k(f(x))) - \E\brack{\bar{\r{y}}(\phi_k(f(\r{x})))}} > \epsilon}\;.
\]
We will do so by bounding this probability with the probability of the ``symmetrized event"
\[
B_{2k,\epsilon} = \set{(xx', yy') \in \mathcal{S}^{2n}: \exists f\in \mathcal{F} \st \abs{\bar{y}(\phi_k(f(x))) - \bar{y}(\phi_k(f(x')))} > \epsilon}\;.
\]
First we bound the probability of $B_{2k, \epsilon}$ and then we relate $B_{2k, \epsilon}$ back to $A_{k, \epsilon}$. To make things slightly more transparent, we write
\begin{align*}
\P^{2n}(B_{2k, \epsilon}) &= \P^{2n}\set{(xx', yy') \in \mathcal{S}^{2n}: \exists f\in \mathcal{F} \st 
\abs{\bar{y}(\phi_k(f(x))) - \bar{y}(\phi_k(f(x')))}> \epsilon} \\[5px]
&= \int_{S^{2n}} \one{\exists f\in \mathcal{F} \st \abs{\bar{y}(\phi_k(f(x))) - \bar{y}(\phi_k(f(x')))}> \epsilon}\; \d{\P^{2n}}\;.
\end{align*}
Now, let $\Lambda_{2n}$ be the subgroup of permutations $\text{Sym}[2n]$ generated by swapping pairs of indices $(i, n+i)$ for $i = 1,\ldots,n$. For $\sigma\in \Lambda_{2n}$, we write 
\[
\sigma(yy') = (\sigma y, \sigma y')\doteq (y_{\sigma_1}, y_{\sigma_2}, \ldots, y_{\sigma_n}, y'_{\sigma_{n+1}}, \ldots, y'_{\sigma_{2n}})\;.
\]
Since the action of $\text{Sym}[2n]$ is measure-preserving with respect to the product measure $\P^{2n}$, the value of the integral above doesn't change if we apply an element $\sigma\in \Lambda_{2n}$ to the indices $[2n]$. In particular, the integral is equal to the expression below for any $\sigma\in \Lambda_{2n}$
\[
\int_{\mathcal{S}^{2n}}\one{\exists f\in \mathcal{F} \st \abs{\overline{\sigma y}(\phi_k(f(x))) - \overline{\sigma y}'(\phi_k(f(x')))} > \epsilon}\;\d{\P^{2n}}\;.
\]
Furthermore, we can average over all permutations in $\Lambda_{2n}$ and still not affect the value of the integral
\[
\int_{\mathcal{S}^{2n}}\frac{1}{|\Lambda_{2n}|}\sum_{\sigma\in \Lambda_{2n}}\one{\exists f\in \mathcal{F} \st \abs{\overline{\sigma y}(\phi_k(f(x))) - \overline{\sigma y}'(\phi_k(f(x')))} > \epsilon}\;\d{\P^{2n}}\;.
\]
For any particular $x$, there are only a finite number of possible $(n,k)$-\textit{order behaviors} $\phi_k(f(x))$ as $f$ ranges over the class $\mathcal{F}$. If we define
\begin{align*}
V_{n,k,x}(\mathcal{F}) = \set{\phi_k(f(x)): f\in \mathcal{F}}\;,
\end{align*}
then we can bound the previous integral as follows:
\begin{align*}
&\int_{\mathcal{S}^{2n}}\frac{1}{|\Lambda_{2n}|}\sum_{\sigma\in \Lambda_{2n}}\one{\exists f\in \mathcal{F} \st \abs{\overline{\sigma y}(\phi_k(f(x))) - \overline{\sigma y}'(\phi_k(f(x')))} > \epsilon}\;\d{\P^{2n}} \\[5px]
&\le \int_{\mathcal{S}^{2n}}\frac{1}{|\Lambda_{2n}|}\sum_{\sigma\in \Lambda_{2n}}\one{\exists C\in V_{n,k,x}(\mathcal{F})\;,\; \exists C'\in V_{n,k,x'}(\mathcal{F}) \st \abs{\overline{\sigma y}(C) - \overline{\sigma y}'(C')} > \epsilon}\;\d{\P^{2n}} \\[5px]
&\le \int_{\mathcal{S}^{2n}}\sum_{\substack{C\in V_{n,k,x}(\mathcal{F})\\ C'\in V_{n,k,x'}(\mathcal{F})}}\paren{\frac{1}{|\Lambda_{2n}|}\sum_{\sigma\in \Lambda_{2n}} \one{ \abs{\overline{\sigma y}(C) - \overline{\sigma y}'(C')} > \epsilon}}\;\d{\P^{2n}}\;.
\end{align*}
The parenthetical expression can be seen to be the probability of a permutation $\sigma\in \Lambda_{2n}$ selected uniformly at random satisfying a particular condition; since $\Lambda_{2n}$ is generated by coordinate-wise swaps, the uniform distribution over $\Lambda_{2n}$ can be decomposed into a product distribution over independent swaps 
\[
\r{\sigma} \overset{d}{=} (\r{s}_1, \ldots, \r{s}_n, \tilde{\r{s}}_1, \ldots, \tilde{\r{s}}_n)
\]
where 
\[
\r{s}_i = \begin{cases}
i & \wp 1/2 \\
n+i & \wp 1/2
\end{cases}
\quad, \quad
\tilde{\r{s}}_i = \begin{cases}
n+i & \text{if } \r{s}_i = i \\
i & \text{if } \r{s}_i = n+i\;.
\end{cases}
\]

are drawn independently for $i = 1,\ldots,n$. Now, we notice that if we ``flip" the value of any $\r{s}_i$ then the value of the random variable $\overline{\r{\sigma} y}(C) - \overline{\r{\sigma} y}'(C')$ changes by at most $2/k$. Furthermore, by symmetry, it's clear that this random variable has mean zero, so, McDiarmid's inequality implies that 
\begin{align*}
\underset{\r{\sigma}\sim \text{unif}(\Lambda_{2n})}{\text{Pr}}\set{\abs{\overline{\r{\sigma} y}(C) - \overline{\r{\sigma} y}'(C')} > \epsilon} &\le 2\exp\paren{-\frac{2\epsilon^2}{n(2/k)^2}} = 2\exp\paren{-\frac{\epsilon^2k^2}{2n}}\;.
\end{align*}
Now we define the $(n,k)$-\textit{order coefficient} to be
\[
\Theta(\mathcal{F}, n, k) = \max_{x\in \mathcal{X}^n} \, |V_{n,k,x}(\mathcal{F})|
\]
and we have that
\begin{align*}
&\int_{\mathcal{S}^{2n}}\sum_{\substack{C\in V_{n,x}(\mathcal{F})\\ C'\in V_{n,x'}(\mathcal{F})}}\paren{\frac{1}{|\Lambda_{2n}|}\sum_{\sigma\in \Lambda_{2n}} \one{ \abs{\overline{\sigma y}(C) - \overline{\sigma y}'(C')} > \epsilon}}\;\d{\P^{2n}}  \\[5px]
& \qquad \qquad \qquad \le   2\cdot \Theta(\mathcal{F}, n, k)^2 \cdot \exp\paren{-\frac{\epsilon^2k^2}{2n}}\;.
\end{align*}
Returning to our original calculation, we have now shown that
\begin{align*}
\P^{2n}(B_{2k, \epsilon}) \le 2\cdot \Theta(\mathcal{F}, n, k)^2 \cdot \exp\paren{-\frac{\epsilon^2k^2}{2n}} \;.
\end{align*}

Our next task is to ``reduce" $\P^n(A_{k, \epsilon})$ to $\P^{2n}(B_{2k, \epsilon})$. Suppose $(x, y) \in \mathcal{S}^n$ falls in $A_{k, \epsilon}$; i.e. $(x,y)$ is such that there exists an $f^* \in \mathcal{F}$ which satisfies 
\[
\abs{\bar{y}(\phi_k(f^*(x))) - \E\brack{\bar{\r{y}}(\phi_k(f^*(\r{x})))}} > \epsilon\;.
\]
Furthermore, suppose $(x', y') \in \mathcal{S}^n$ is such that $\abs{\bar{y}(\phi_k(f^*(x'))) - \E\brack{\bar{\r{y}}(\phi_k(f^*(\r{x})))}} \le \epsilon/2$. Then by the triangle inequality, we have that
\begin{align*}
\abs{\bar{y}(\phi_k(f(x))) - \bar{y}(\phi_k(f(x')))} &\ge \abs{\bar{y}(\phi_k(f^*(x))) - \E\brack{\bar{\r{y}}(\phi_k(f^*(\r{x})))}} \\
&\hspace{2cm} - \abs{\bar{y}(\phi_k(f^*(x'))) - \E\brack{\bar{\r{y}}(\phi_k(f^*(\r{x})))}} \\[5px]
&> \epsilon - \epsilon/2 \\[5px]
&= \epsilon/2\;.
\end{align*}
This implies that 
\begin{align*}
\{(x, y) \in \mathcal{S}^n: \exists f\in \mathcal{F} \st &\abs{\bar{y}(\phi_k(f(x))) - \E\brack{\bar{\r{y}}(\phi_k(f(\r{x})))}} > \epsilon \;, \; \\[5px]
 &\abs{\bar{y}(\phi_k(f(x'))) - \E\brack{\bar{\r{y}}(\phi_k(f(\r{x})))}} \le \epsilon/2
\} \subseteq
B_{2k, \epsilon/2} \tag{*}\;.
\end{align*}
Since the random variable $\bar{y}(\phi_k(f(x')))$ satisfies the bounded-differences condition with parameter $1/k$, an application of McDiarmid's inequality yields that for any $f$
\begin{align*}
\P^n\set{(x', y') \in \mathcal{S}^n: \abs{\bar{y}(\phi_k(f(x'))) - \E\brack{\bar{\r{y}}(\phi_k(f(\r{x})))}} > \epsilon/2} &\le 2 \exp\paren{\frac{-2(\epsilon/2)^2k^2}{n}} \\[5px]
&= 2 \exp\paren{-\frac{\epsilon^2k^2}{2n}}
\end{align*}
and therefore, 
\[
\P^n\set{(x', y') \in \mathcal{S}^n: \abs{\bar{y}(\phi_k(f(x'))) - \E\brack{\bar{\r{y}}(\phi_k(f(\r{x})))}} \le \epsilon/2} > 1 - 2 \exp\paren{-\frac{\epsilon^2k^2}{2n}}
\]
as long as $\epsilon^2k^2/2n > \log(2)$. By taking probabilities on both sides of $(*)$ and applying the Fubini theorem, we see that 
\begin{align*}
\P^n(A_{k,\epsilon})  &\le \frac{1}{1 - 2\exp\paren{-\epsilon^2k^2/2n}}\cdot \P^{2n}(B_{2k, \epsilon/2}) \\[5px]
&\le \Theta(\mathcal{F}, n, k)^2 \cdot \frac{2\exp\paren{-\epsilon^2k^2/2n}}{1 - 2\exp\paren{-\epsilon^2k^2/2n}} \\[5px]
&\le 4\cdot \Theta(\mathcal{F}, n, k)^2\cdot \exp\paren{-\frac{\epsilon^2k^2}{2n}}
\end{align*}

where in the last line we used the fact that $2e^{-t}/(1-2e^{-t}) \le 4e^{-t}$ when $t > \log(2)$ and $4e^{-t} \ge 1$ when $t \le \log(2)$; this coincides with the region for which the inequality in the second line holds nontrivially. Setting the right hand side equal to $\delta/2n$, solving for $\epsilon$, taking a union bound over $k = 1,\ldots,n$, and applying Lemma 1 for the bias terms, yields the first bound in the theorem. The analogous result for negative predictive value is derived in exactly the same fashion. \qed

\end{document}